# Some Novel Applications of Explanation-Based Learning to Parsing Lexicalized Tree-Adjoining Grammars*

**B. Srinivas** and **Aravind K. Joshi**
Department of Computer and Information Science
University of Pennsylvania
Philadelphia, PA 19104, USA
{srini, joshi}@linc.cis.upenn.edu

## Abstract

In this paper we present some novel applications of Explanation-Based Learning (EBL) technique to parsing Lexicalized Tree-Adjoining grammars. The novel aspects are (a) immediate generalization of parses in the training set, (b) generalization over recursive structures and (c) representation of generalized parses as Finite State Transducers. A highly impoverished parser called a "stapler" has also been introduced. We present experimental results using EBL for different corpora and architectures to show the effectiveness of our approach.

## 1 Introduction

In this paper we present some novel applications of the so-called Explanation-Based Learning technique (EBL) to parsing Lexicalized Tree-Adjoining grammars (LTAG). EBL techniques were originally introduced in the AI literature by (Mitchell et al., 1986; Minton, 1988; van Harmelen and Bundy, 1988). The main idea of EBL is to keep track of problems solved in the past and to replay those solutions to solve new but somewhat similar problems in the future. Although put in these general terms the approach sounds attractive, it is by no means clear that EBL will actually improve the performance of the system using it, an aspect which is of great interest to us here.

Rayner (1988) was the first to investigate this technique in the context of natural language parsing. Seen as an EBL problem, the parse of a single sentence represents an *explanation* of why the sentence is a part of the language defined by the grammar. Parsing new sentences amounts to finding analogous explanations from the training sentences. As a special case of EBL, Samuelsson and Rayner (1991) specialize a grammar for the ATIS domain by storing chunks of the parse trees present in a treebank of parsed examples. The idea is to reparse the training examples by letting the parse tree drive the rule expansion process and halting the expansion of a specialized rule if the current node meets a 'tree-cutting' criteria. However, the problem of specifying an optimal 'tree-cutting' criteria was not addressed in this work. Samuelsson (1994) used the information-theoretic measure of entropy to derive the appropriate sized tree chunks automatically. Neumann (1994) also attempts to specialize a grammar given a training corpus of parsed examples by generalizing the parse for each sentence and storing the generalized phrasal derivations under a suitable index.

Although our work can be considered to be in this general direction, it is distinct in that it exploits some of the key properties of LTAG to (a) achieve an *immediate* generalization of parses in the training set of sentences, (b) achieve an additional level of generalization of the parses in the training set, thereby dealing with test sentences which are not necessarily of the same length as the training sentences and (c) represent the set of generalized parses as a finite state transducer (FST), which is the first such use of FST in the context of EBL, to the best of our knowledge. Later in the paper, we will make some additional comments on the relationship between our approach and some of the earlier approaches.

In addition to these special aspects of our work, we will present experimental results evaluating the effectiveness of our approach on more than one kind of corpus. We also introduce a device called a "stapler", a considerably impoverished parser, whose only job is to do term unification and compute alternate attachments for modifiers. We achieve substantial speed-up by the use of "stapler" in combination with the output of the FST.

The paper is organized as follows. In Section 2 we provide a brief introduction to LTAG with the help of an example. In Section 3 we discuss our approach to using EBL and the advantages provided

*This work was partially supported by ARO grant DAAL03-89-0031, ARPA grant N00014-90-J-1863, NSF STC grant DIR-8920230, and Ben Franklin Partnership Program (PA) grant 93S.3078C-6

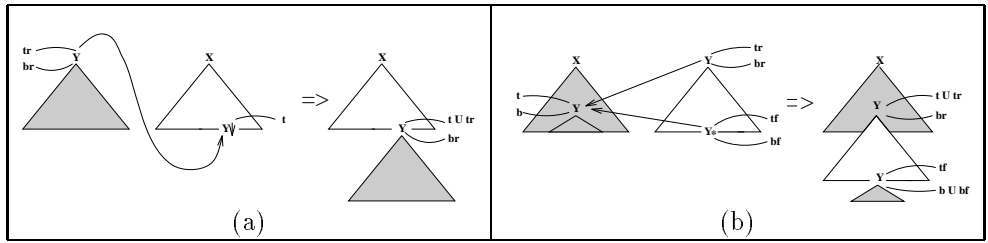

Figure 1: Substitution and Adjunction in LTAG

by LTAG. The FST representation used for EBL is illustrated in Section 4. In Section 5 we present the "stapler" in some detail. The results of some of the experiments based on our approach are presented in Section 6. In Section 7 we discuss the relevance of our approach to other lexicalized grammars. In Section 8 we conclude with some directions for future work.

## 2 Lexicalized Tree-Adjoining Grammar

Lexicalized Tree-Adjoining Grammar (LTAG) (Schabes et al., 1988; Schabes, 1990) consists of ELEMENTARY TREES, with each elementary tree having a lexical item (anchor) on its frontier. An elementary tree serves as a complex description of the anchor and provides a domain of locality over which the anchor can specify syntactic and semantic (predicate-argument) constraints. Elementary trees are of two kinds – (a) INITIAL TREES and (b) AUXILIARY TREES.

Nodes on the frontier of initial trees are marked as substitution sites by a '↓'. Exactly one node on the frontier of an auxiliary tree, whose label matches the label of the root of the tree, is marked as a foot node by a '∗'; the other nodes on the frontier of an auxiliary tree are marked as substitution sites. Elementary trees are combined by **Substitution** and **Adjunction** operations.

Each node of an elementary tree is associated with the top and the bottom feature structures (FS). The bottom FS contains information relating to the subtree rooted at the node, and the top FS contains information relating to the supertree at that node.[1] The features may get their values from three different sources such as the morphology of anchor, the structure of the tree itself, or by unification during the derivation process. FS are manipulated by substitution and adjunction as shown in Figure 1.

The initial trees ($\alpha$s) and auxiliary trees ($\beta$s) for the sentence *show me the flights from Boston to Philadelphia* are shown in Figure 2. Due to the limited space, we have shown only the features on the $\alpha_1$ tree. The result of combining the elementary trees shown in Figure 2 is the **derived tree**, shown in Figure 2(a). The process of combining the elementary trees to yield a parse of the sentence is represented by the **derivation tree**, shown in Figure 2(b). The nodes of the derivation tree are the tree names that are anchored by the appropriate lexical items. The combining operation is indicated by the nature of the arcs–broken line for substitution and bold line for adjunction–while the address of the operation is indicated as part of the node label. The derivation tree can also be interpreted as a *dependency tree*[2] with unlabeled arcs between words of the sentence as shown in Figure 2(c).

Elementary trees of LTAG are the domains for specifying dependencies. Recursive structures are specified via the auxiliary trees. The three aspects of LTAG – (a) lexicalization, (b) extended domain of locality and (c) factoring of recursion, provide a natural means for generalization during the EBL process.

## 3 Overview of our approach to using EBL

We are pursuing the EBL approach in the context of a wide-coverage grammar development system called XTAG (Doran et al., 1994). The XTAG system consists of a morphological analyzer, a part-of-speech tagger, a wide-coverage LTAG English grammar, a predictive left-to-right Early-style parser for LTAG (Schabes, 1990) and an X-windows interface for grammar development (Paroubek et al., 1992). Figure 3 shows a flowchart of the XTAG system. The input sentence is subjected to morphological analysis and is parts-of-speech tagged before being sent to the parser. The parser retrieves the elementary trees that the words of the sentence anchor and combines them by adjunction and substitution operations to derive a parse of the sentence.

Given this context, the training phase of the EBL process involves generalizing the derivation trees generated by XTAG for a training sentence and storing these generalized parses in the generalized parse database under an index computed from the morphological features of the sentence. The application phase of EBL is shown in the flowchart in Figure 4. An index using the morphological features of the words in the input sentence is computed. Using this index, a set of generalized parses is retrieved from the generalized parse database created in the training phase. If the retrieval fails to yield any generalized parse then the input sentence is parsed using the full parser. However, if the retrieval succeeds then the generalized parses are input to the "stapler". Section 5 provides a description of the "stapler".

---
[1] Nodes marked for substitution are associated with only the top FS.

[2] There are some differences between derivation trees and conventional dependency trees. However we will not discuss these differences in this paper as they are not relevant to the present work.

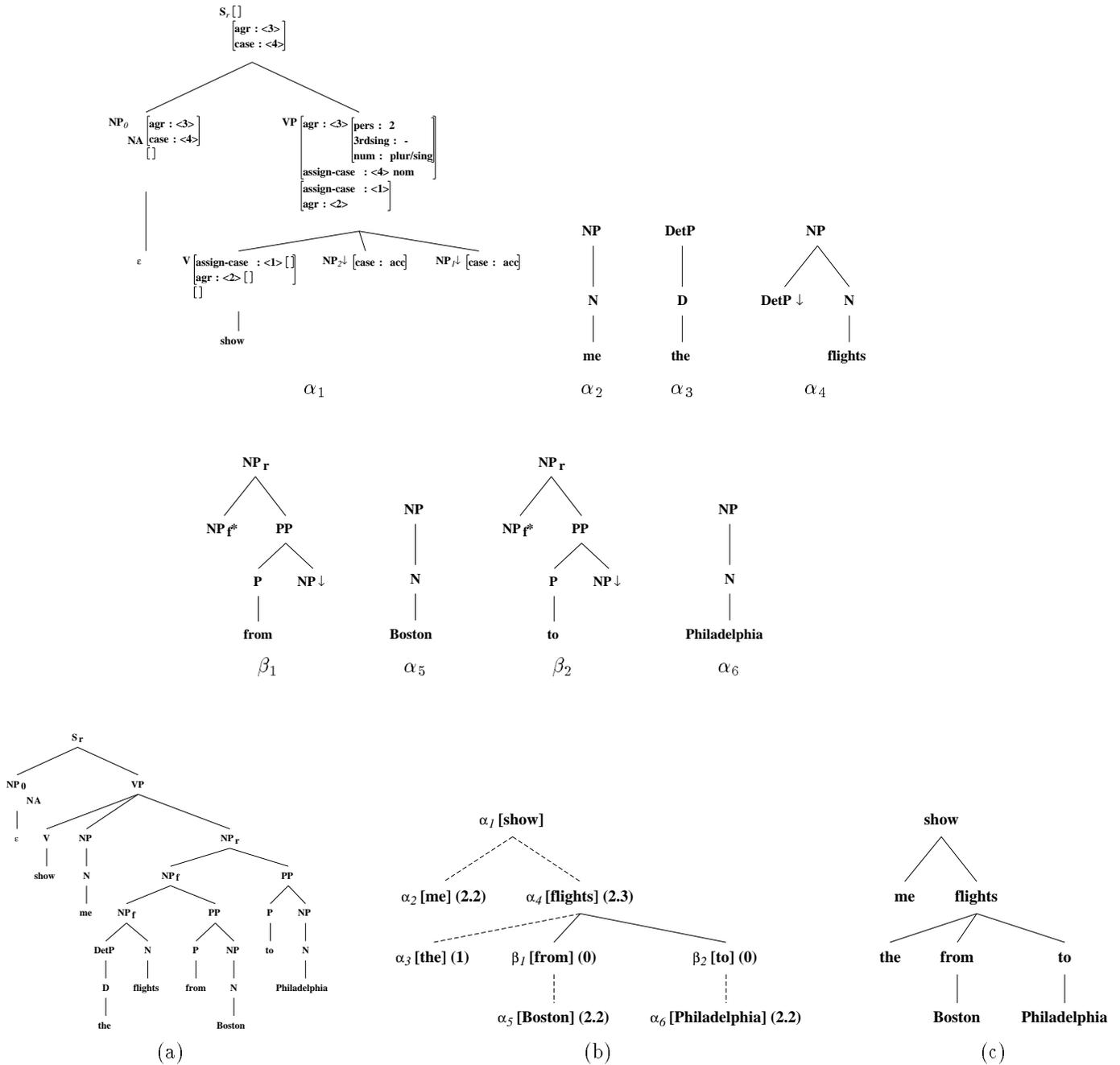

Figure 2: ($\alpha$s and $\beta$s) Elementary trees, (a) Derived Tree, (b) Derivation Tree, and (c) Dependency tree for the sentence: *show me the flights from Boston to Philadelphia*.

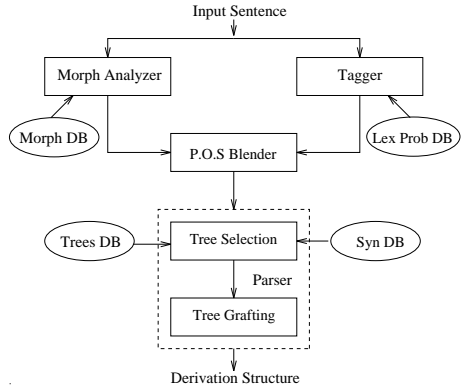

Figure 3: **Flowchart of the XTAG system**

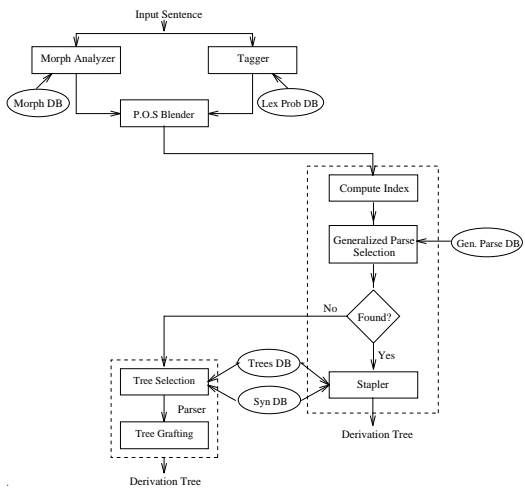

Figure 4: **Flowchart of the XTAG system with the EBL component**

### 3.1 Implications of LTAG representation for EBL

An LTAG parse of a sentence can be seen as a sequence of elementary trees associated with the lexical items of the sentence along with substitution and adjunction links among the elementary trees. Also, the feature values in the feature structures of each node of every elementary tree are instantiated by the parsing process. Given an LTAG parse, the generalization of the parse is truly *immediate* in that a generalized parse is obtained by (a) uninstantiating the particular lexical items that anchor the individual elementary trees in the parse and (b) uninstantiating the feature values contributed by the morphology of the anchor and the derivation process. This type of generalization is called *feature-generalization*.

In other EBL approaches (Rayner, 1988; Neumann, 1994; Samuelsson, 1994) it is necessary to walk up and down the parse tree to determine the appropriate subtrees to generalize on and to suppress the feature values. In our approach, the process of generalization is *immediate*, once we have the output of the parser, since the elementary trees anchored by the words of the sentence define the subtrees of the parse for generalization. Replacing the elementary trees with unistantiated feature values is all that is needed to achieve this generalization.

The generalized parse of a sentence is stored indexed on the part-of-speech (POS) sequence of the training sentence. In the application phase, the POS sequence of the input sentence is used to retrieve a generalized parse(s) which is then instantiated with the features of the sentence. This method of retrieving a generalized parse allows for parsing of sentences of the same lengths and the same POS sequence as those in the training corpus. However, in our approach there is another generalization that falls out of the LTAG representation which allows for flexible matching of the index to allow the system to parse sentences that are not necessarily of the same length as any sentence in the training corpus.

Auxiliary trees in LTAG represent recursive structures. So if there is an auxiliary tree that is used in an LTAG parse, then that tree with the trees for its arguments can be repeated any number of times, or possibly omitted altogether, to get parses of sentences that differ from the sentences of the training corpus only in the number of modifiers. This type of generalization is called *modifier-generalization*. This type of generalization is not possible in other EBL approaches.

This implies that the POS sequence covered by the auxiliary tree and its arguments can be repeated zero or more times. As a result, the index of a generalized parse of a sentence with modifiers is no longer a string but a regular expression pattern on the POS sequence and retrieval of a generalized parse involves regular expression pattern matching on the indices. If, for example, the training example was

(1) Show/V me/N the/D flights/N from/P Boston/N to/P Philadelphia/N.

then, the index of this sentence is

(2) V N D N (P N)*

since the two prepositions in the parse of this sentence would anchor (the same) auxiliary trees.

The most efficient method of performing regular expression pattern matching is to construct a *finite state machine* for each of the stored patterns and then traverse the machine using the given test pattern. If the machine reaches the final state, then the test pattern matches one of the stored patterns.

Given that the index of a test sentence matches one of the indices from the training phase, the generalized parse retrieved will be a parse of the test sentence, modulo the modifiers. For example, if the test sentence, tagged appropriately, is

(3) Show/V me/N the/D flights/N from/P Boston/N to/P Philadelphia/N on/P Monday/N.

then, although the index of the test sentence matches the index of the training sentence, the generalized parse retrieved needs to be augmented to accommodate the additional modifier.

To accommodate the additional modifiers that may be present in the test sentences, we need to provide a mechanism that assigns the additional modifiers and their arguments the following:

1. The elementary trees that they anchor and
2. The substitution and adjunction links to the trees they substitute or adjoin into.

We assume that the additional modifiers along with their arguments would be assigned the same elementary trees and the same substitution and adjunction links as were assigned to the modifier and its arguments of the training example. This, of course, means that we may not get all the possible attachments of the modifiers at this time. (but see the discussion of the "stapler" Section 5.)

## 4 FST Representation

The representation in Figure 6 combines the generalized parse with the POS sequence (regular expression) that it is indexed by. The idea is to annotate each of the finite state arcs of the regular expression matcher with the elementary tree associated with that POS and also indicate which elementary tree it would be adjoined or substituted into. This results in a *Finite State Transducer (FST)* representation, illustrated by the example below. Consider the sentence (4) with the derivation tree in Figure 5.

(4) show me the flights from Boston to Philadelphia.

An alternate representation of the derivation tree that is similar to the *dependency representation*, is to associate with each word a tuple (this_tree, head_word, head_tree, number). The description of the tuple components is given in Table 1.

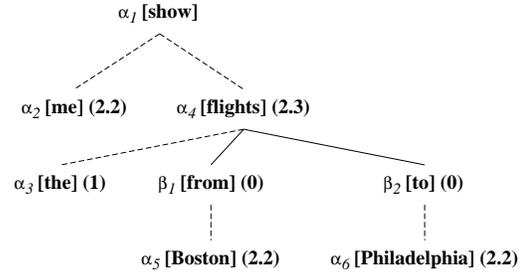

Figure 5: Derivation Tree for the sentence: *show me the flights from Boston to Philadelphia*

| | | |
|---|---|---|
| this_tree | : | the elementary tree that the word anchors |
| head_word | : | the word on which the current word is dependent on; "–" if the current word does not depend on any other word. |
| head_tree | : | the tree anchored by the head word; "–" if the current word does not depend on any other word. |
| number | : | a signed number that indicates the direction and the ordinal position of the particular head elementary tree from the position of the current word *OR* |
| | : | an unsigned number that indicates the Gorn-address (i.e., the node address) in the derivation tree to which the word attaches *OR* |
| | : | "–" if the current word does not depend on any other word. |

Table 1: Description of the tuple components

Following this notation, the derivation tree in Figure 5 (without the addresses of operations) is represented as in (5).

(5)
$$\begin{array}{ll}
\text{show}/(\alpha_1, -, -, -) & \text{me}/(\alpha_2, \text{show}, \alpha_1, -1) \\
\text{the}/(\alpha_3, \text{flights}, \alpha_4, +1) & \text{flights}/(\alpha_4, \text{show}, \alpha_1, -1) \\
\text{from}/(\beta_1, \text{flights}, \alpha_4, 2) & \text{Boston}/(\alpha_5, \text{from}, \beta_1, -1) \\
\text{to}/(\beta_2, \text{flights}, \alpha_4, 2) & \text{Philadelphia}/(\alpha_6, \text{to}, \beta_2, -1)
\end{array}$$

Generalization of this derivation tree results in the representation in (6).

(6)
$$\begin{array}{ll}
\text{V}/(\alpha_1, -, -, -) & \text{N}/(\alpha_2, \text{V}, \alpha_1, -1) \\
\text{D}/(\alpha_3, \text{N}, \alpha_4, +1) & \text{N}/(\alpha_4, \text{V}, \alpha_1, -1) \\
(\text{P}/(\beta_1, \text{N}, \alpha_4, 2) & \text{N}/(\alpha_5, \text{P}, \beta, -1))^* \\
(\text{P}/(\beta_2, \text{N}, \alpha_4, 2) & \text{N}/(\alpha_6, \text{P}, \beta, -1))^*
\end{array}$$

After generalization, the trees $\beta_1$ and $\beta_2$ are no longer distinct so we denote them by $\beta$. The trees $\alpha_5$ and $\alpha_6$ are also no longer distinct, so we denote them by $\alpha$. With this change in notation, the two

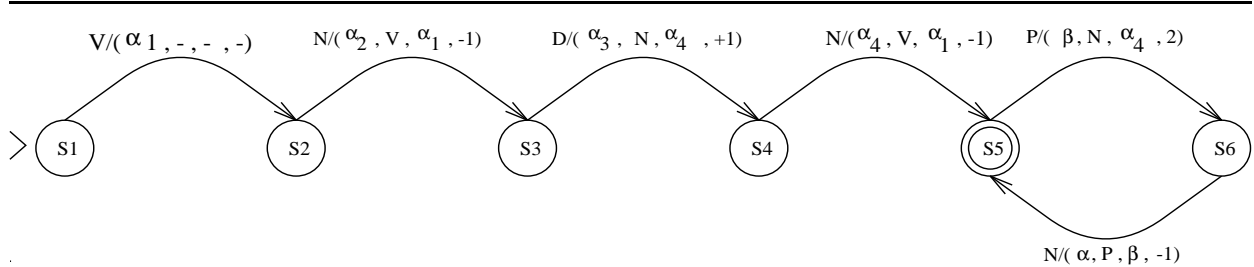

Figure 6: Finite State Transducer Representation for the sentences: *show me the flights from Boston to Philadelphia, show me the flights from Boston to Philadelphia on Monday, ...*

Kleene star regular expressions in (6) can be merged into one, and the resulting representation is (7)

(7) $\quad \begin{array}{ll} V/(\alpha_1, -, -, -) & N/(\alpha_2, V, \alpha_1, -1) \\ D/(\alpha_3, N, \alpha_4, +1) & N/(\alpha_4, V, \alpha_1, -1) \\ (P/(\beta, N, \alpha_4, 2) & N/(\alpha, P, \beta, -1) )^* \end{array}$

which can be seen as a path in an FST as in Figure 6.

This FST representation is possible due to the lexicalized nature of the elementary trees. This representation makes a distinction between dependencies between modifiers and complements. The number in the tuple associated with each word is a signed number if a complement dependency is being expressed and is an unsigned number if a modifier dependency is being expressed.[3]

## 5 Stapler

In this section, we introduce a device called "stapler", a very impoverished parser that takes as input the result of the EBL lookup and returns the parse(s) for the sentence. The output of the EBL lookup is a sequence of elementary trees annotated with dependency links – an *almost parse*. To construct a complete parse, the "stapler" performs the following tasks:

- Identify the nature of link: The dependency links in the *almost parse* are to be distinguished as either substitution links or adjunction links. This task is extremely straightforward since the types (initial or auxiliary) of the elementary trees a dependency link connects identifies the nature of the link.

- Modifier Attachment: The EBL lookup is not guaranteed to output all possible modifier-head dependencies for a give input, since the modifier-generalization assigns the same modifier-head link, as was in the training example, to all the additional modifiers. So it is the task of the stapler to compute all the alternate attachments for modifiers.

- Address of Operation: The substitution and adjunction links are to be assigned a node address to indicate the location of the operation. The "stapler" assigns this using the structure of the elementary trees that the words anchor and their linear order in the sentence.

- Feature Instantiation: The values of the features on the nodes of the elementary trees are to be instantiated by a process of unification. Since the features in LTAGs are finite-valued and only features within an elementary tree can be co-indexed, the "stapler" performs term-unification to instantiate the features.

## 6 Experiments and Results

We now present experimental results from two different sets of experiments performed to show the effectiveness of our approach. The first set of experiments, (Experiments 1(a) through 1(c)), are intended to measure the coverage of the FST representation of the parses of sentences from a range of corpora (ATIS, IBM-Manual and Alvey). The results of these experiments provide a measure of repetitiveness of patterns as described in this paper, at the sentence level, in each of these corpora.

**Experiment 1(a):** The details of the experiment with the ATIS corpus are as follows. A total of 465 sentences, average length of 10 words per sentence, which had been completely parsed by the XTAG system were randomly divided into two sets, a training set of 365 sentences and a test set of 100 sentences, using a random number generator. For each of the training sentences, the parses were ranked using heuristics[4] (Srinivas et al., 1994) and the top three derivations were generalized and stored as an FST. The FST was tested for retrieval of a generalized parse for each of the test sentences that were pretagged with the correct POS sequence (In Experiment 2, we make use of the POS tagger to do the tagging). When a match is found, the output of the EBL component is a generalized parse that associates with each word the elementary tree that

---

[3]In a complement auxiliary tree the anchor subcategorizes for the foot node, which is not the case for a modifier auxiliary tree.

[4]We are not using stochastic LTAGs. For work on Stochastic LTAGs see (Resnik, 1992; Schabes, 1992).

| Corpus | Size of Training set | # of States | % Coverage | Response Time (secs) |
|---|---|---|---|---|
| ATIS | 365 | 6000 | 80% | 1.00 sec/sent |
| IBM | 1100 | 21000 | 40% | 4.00 sec/sent |
| Alvey | 80 | 500 | 50% | 0.20 sec/NP |

Table 2: Coverage and Retrieval times for various corpora

it anchors and the elementary tree into which it adjoins or substitutes into – an *almost parse*.[5]

**Experiment 1(b) and 1(c):** Similar experiments were conducted using the IBM-manual corpus and a set of noun definitions from the LDOCE dictionary that were used as the Alvey test set (Carroll, 1993).

Results of these experiments are summarized in Table 2. The size of the FST obtained for each of the corpora, the coverage of the FST and the traversal time per input are shown in this table. The coverage of the FST is the number of inputs that were assigned a correct generalized parse among the parses retrieved by traversing the FST.

Since these experiments measure the performance of the EBL component on various corpora we will refer to these results as the 'EBL-Lookup times'.

The second set of experiments measure the performance improvement obtained by using EBL within the XTAG system on the ATIS corpus. The performance was measured on the same set of 100 sentences that was used as test data in Experiment 1(a). The FST constructed from the generalized parses of the 365 ATIS sentences used in experiment 1(a) has been used in this experiment as well.

**Experiment 2(a):** The performance of XTAG on the 100 sentences is shown in the first row of Table 3. The coverage represents the percentage of sentences that were assigned a parse.

**Experiment 2(b):** This experiment is similar to Experiment 1(a). It attempts to measure the coverage and response times for retrieving a generalized parse from the FST. The results are shown in the second row of Table 3. The difference in the response times between this experiment and Experiment 1(a) is due to the fact that we have included here the times for morphological analysis and the POS tagging of the test sentence. As before, 80% of the sentences were assigned a generalized parse. However, the speedup when compared to the XTAG system is a factor of about 60.

**Experiment 2(c):** The setup for this experiment is shown in Figure 7. The *almost parse* from the EBL lookup is input to the full parser of the XTAG system. The full parser does not take advantage of the dependency information present in the *almost parse*, however it benefits from the elementary tree assignment to the words in it. This information helps the full parser, by reducing the ambiguity of assigning a correct elementary tree sequence for the words of the sentence. The speed up shown in the third row of Table 3 is entirely due to this ambiguity reduction. If the EBL lookup fails to retrieve a parse, which happens for 20% of the sentences, then the tree assignment ambiguity is not reduced and the full parser parses with all the trees for the words of the sentence. The drop in coverage is due to the fact that for 10% of the sentences, the generalized parse retrieved could not be instantiated to the features of the sentence.

---
[5]See (Joshi and Srinivas, 1994) for the role of almost parse in supertag disambiguation.

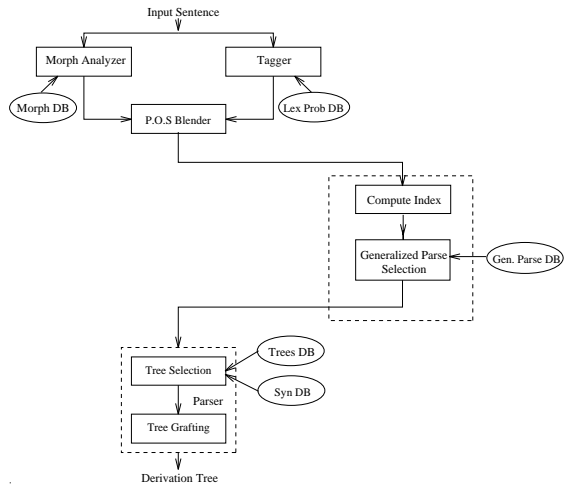

Figure 7: **System Setup for Experiment 2(c).**

| System | Coverage % | Average time (in secs) |
|---|---|---|
| XTAG | 100% | 125.18 |
| EBL lookup | 80% | 1.78 |
| EBL+XTAG parser | 90% | 62.93 |
| EBL+Stapler | 70% | 8.00 |

Table 3: Performance comparison of XTAG with and without EBL component

**Experiment 2(d):** The setup for this experiment is shown in Figure 4. In this experiment, the almost parse resulting from the EBL lookup is input to the "stapler" that generates all possible modifier attachments and performs term unification thus generating all the derivation trees. The "stapler" uses both the elementary tree assignment information and the dependency information present in the almost parse and speeds up the performance even further, by a factor of about 15 with further decrease in coverage by 10% due to the same reason as mentioned in Experiment 2(c). However the cov-erage of this system is limited by the coverage of the EBL lookup. The results of this experiment are shown in the fourth row of Table 3.

## 7 Relevance to other lexicalized grammars

Some aspects of our approach can be extended to other lexicalized grammars, in particular to categorial grammars (e.g. Combinatory Categorial Grammar (CCG) (Steedman, 1987)). Since in a categorial grammar the category for a lexical item includes its arguments, the process of generalization of the parse can also be *immediate* in the same sense of our approach. The generalization over recursive structures in a categorial grammar, however, will require further annotations of the proof trees in order to identify the 'anchor' of a recursive structure. If a lexical item corresponds to a potential recursive structure then it will be necessary to encode this information by making the result part of the functor to be $X \to X$. Further annotation of the proof tree will be required to keep track of dependencies in order to represent the generalized parse as an FST.

## 8 Conclusion

In this paper, we have presented some novel applications of EBL technique to parsing LTAG. We have also introduced a highly impoverished parser called the "stapler" that in conjunction with the EBL results in a speed up of a factor of about 15 over a system without the EBL component. To show the effectiveness of our approach we have also discussed the performance of EBL on different corpora, and different architectures.

As part of the future work we will extend our approach to corpora with fewer repetitive sentence patterns. We propose to do this by generalizing at the phrasal level instead of at the sentence level.